\documentclass[conference]{IEEEtran}
\IEEEoverridecommandlockouts
\usepackage{cite}
\usepackage{amsmath,amssymb,amsfonts}
\usepackage{algorithmic}
\usepackage{graphicx}
\usepackage{textcomp}
\usepackage{xcolor}
\usepackage{booktabs}
\usepackage[table]{xcolor}
\usepackage{multirow}
\usepackage{adjustbox}
\usepackage{verbatim}
\usepackage{fancyvrb}
\usepackage{fvextra}
\usepackage{listings}
\def\BibTeX{{\rm B\kern-.05em{\sc i\kern-.025em b}\kern-.08em
    T\kern-.1667em\lower.7ex\hbox{E}\kern-.125emX}}
    
\begin{document}

\title{Multi-LLM Collaborative MRI Report Generation for Visual Instruction Tuning in Brain Oncology
    {\footnotesize} \thanks{* Equal Contribution}
    {\footnotesize} \thanks{$\dagger$ Corresponding Author}
}

\author{
    \IEEEauthorblockN{
        Sinyoung Ra \textsuperscript{1,*},
        Jonghun Kim\textsuperscript{2,*},
        Hyunjin Park\textsuperscript{1,2,$\dagger$}
    }
    \IEEEauthorblockA{\textsuperscript{1}Department of Artificial Intelligence, Sungkyunkwan University, Suwon, Republic of Korea}
    \IEEEauthorblockA{\textsuperscript{2}Department of Electrical and Computer Engineering, Sungkyunkwan University, Suwon, Republic of Korea}
    \{nsy0527, iproj2, hyunjinp\}@skku.edu 
}

\maketitle

\begin{abstract}
Recent advances in large language models (LLMs) and their extension to vision-language models (VLMs) have made it easier to combine text and images for tasks such as report generation. Existing VLMs in medicine typically focus on 2D images (chest X-rays), and their extension to 3D imaging has been difficult because of the lack of paired 3D imaging-text data. Thus, we introduce a new method for creating a 3D image-text dataset for brain oncology using 3D MRI scans of glioma and meningioma cases. We use a cooperative system in which several LLMs work together to generate and check reports, ensuring that they are accurate and clear. By leveraging the new 3D MRI-text dataset, we further build a VLM that converts MRI scans into tokens and aligns them with text instructions. Our VLM performed better in report generation and visual question answering tasks than other 2D and 3D methods. Our method not only improves the quality of reports but also helps with better diagnosis and treatment in brain oncology.


% Authors must provide keywords and are not allowed to remove this Keyword section.

\end{abstract}
\section{Introduction}

Recent advancements in large language models (LLMs) have significantly improved natural language understanding and generation in various domains \cite{zhao2024surveylargelanguagemodels,ren2024advancements,ra2025enhancing}. As an extension, multimodal LLMs, particularly vision-language models (VLMs), have emerged as powerful tools for integrating visual and textual information, enabling applications such as image captioning, report generation, and visual question answering (VQA) \cite{liu2023visual,li2022blip,kim2026visual2}. These methods have been applied in the medical domain, where VLMs have demonstrated notable success in analyzing medical images alongside corresponding textual annotations, enhancing automated clinical decision support and medical documentation \cite{li2023llavamed,lee2023vision,xraygpt,kim2026visual}.

Despite these advances, existing medical VLMs have primarily focused on 2D modalities, particularly chest X-ray datasets, because of their large-scale availability and standardized reporting structures \cite{johnson2019mimic,irvin2019chexpert,kim2025privacy}. Crucially, medical imaging relies heavily on 3D modalities, such as magnetic resonance imaging (MRI) and computed tomography (CT), which provide significantly richer anatomical and pathological information than 2D images. Thus, the development of VLMs capable of understanding 3D medical images is essential for advancing automated diagnosis and clinical decision support. However, extending VLMs to 3D medical imaging presents several challenges. Unlike X-ray datasets, publicly available large-scale image-text paired datasets for 3D medical imaging modalities remain limited. The lack of comprehensive 3D image-text datasets significantly hampers progress in medical VLM research because existing models struggle to generalize beyond 2D image analysis. Addressing this gap is critical for enabling AI-driven interpretation of medical images across a broader range of clinical applications.

\begin{figure}[t]
    \centering
    \includegraphics[width=\columnwidth]{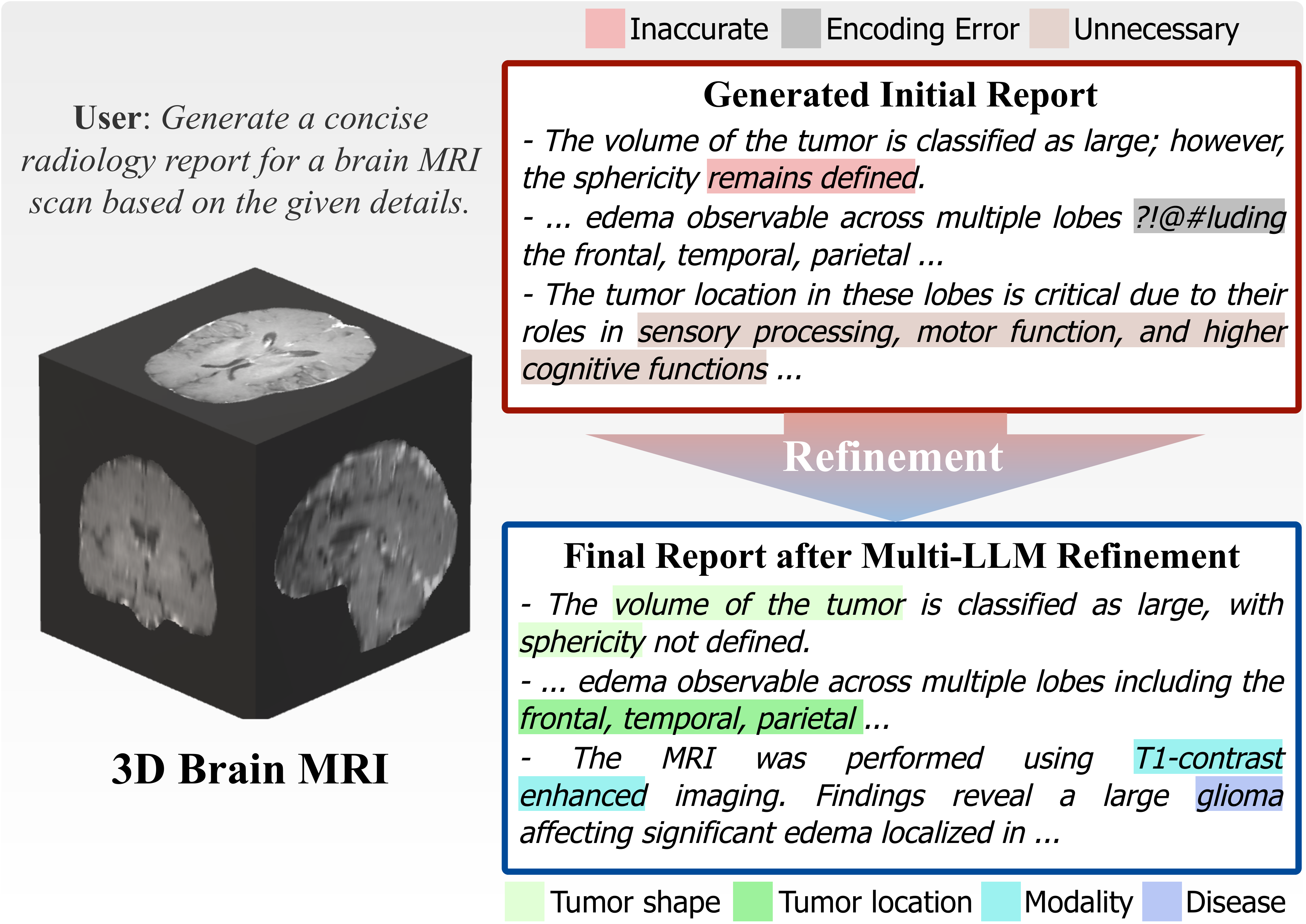}
    \caption{\textbf{Example of Report Generation}. The initial report (red box) is imprecise, includes unnecessary details, and produces hallucinations. Our Multi-LLM refinement process removes unnecessary information and captures detailed image information. The refined report (blue box) accurately generates the essential MRI details.}
    \label{fig1}
\end{figure}

To address this gap, we introduce a novel methodology for generating 3D image-text paired datasets specifically for brain oncology, leveraging glioma and meningioma cases. In contrast to previous works that relied on manual annotations or small-scale datasets, our method facilitates automated MRI report generation using LLMs. To ensure high-quality textual descriptions, we propose a multi-LLM cooperative refinement approach in which multiple LLMs review and verify reports to improve accuracy and consistency. This feedback from multiple LLMs reduces errors and improves the reliability of the generated reports (Fig. \ref{fig1}). Previous studies often relied on clinical experts' validation to ensure the accuracy and relevance of automatically generated radiology reports \cite{tanno2025collaboration}. However, such clinical review processes present significant challenges in practice due to limited expert availability, high costs, and subjective assessment. In response, our approach is specifically motivated by the need to build robust and scalable report generation pipelines that can function without direct clinician involvement. By leveraging collaborative evaluation across multiple independent LLMs, we aim to approximate expert review and establish an automated quality assurance mechanism. Notably, incorporating multi-LLM collaboration in the review process helps to significantly reduce hallucinations and factual inconsistencies \cite{sng2024novel, mousavi2023enhancing}. This shift not only enables large-scale data curation in resource-constrained environments but also opens new directions for validating medical artificial intelligence (AI) outputs when clinical expert access is infeasible or impractical. Our framework has the potential to address a key challenge in previous research: the dependency on costly and time-consuming expert validation, making high-quality report generation more accessible for broad medical AI applications.

Furthermore, we develop a VLM trained on this new dataset by tokenizing MRI images with a VQ-GAN  \cite{esser2021taming} encoder and using the tokens for LLM tuning, demonstrating the new dataset’s utility. This approach enables feature extraction and alignment between the visual and textual modalities, optimizing the model’s ability to generate clinically meaningful reports for MRI. Our study shows that multi-LLM cooperative report generation, combined with VLM-based image analysis, goes beyond traditional 2D imaging and paves the way for more robust multimodal models in brain oncology.

\begin{figure}[t]
    \centering
    \includegraphics[width=\columnwidth]{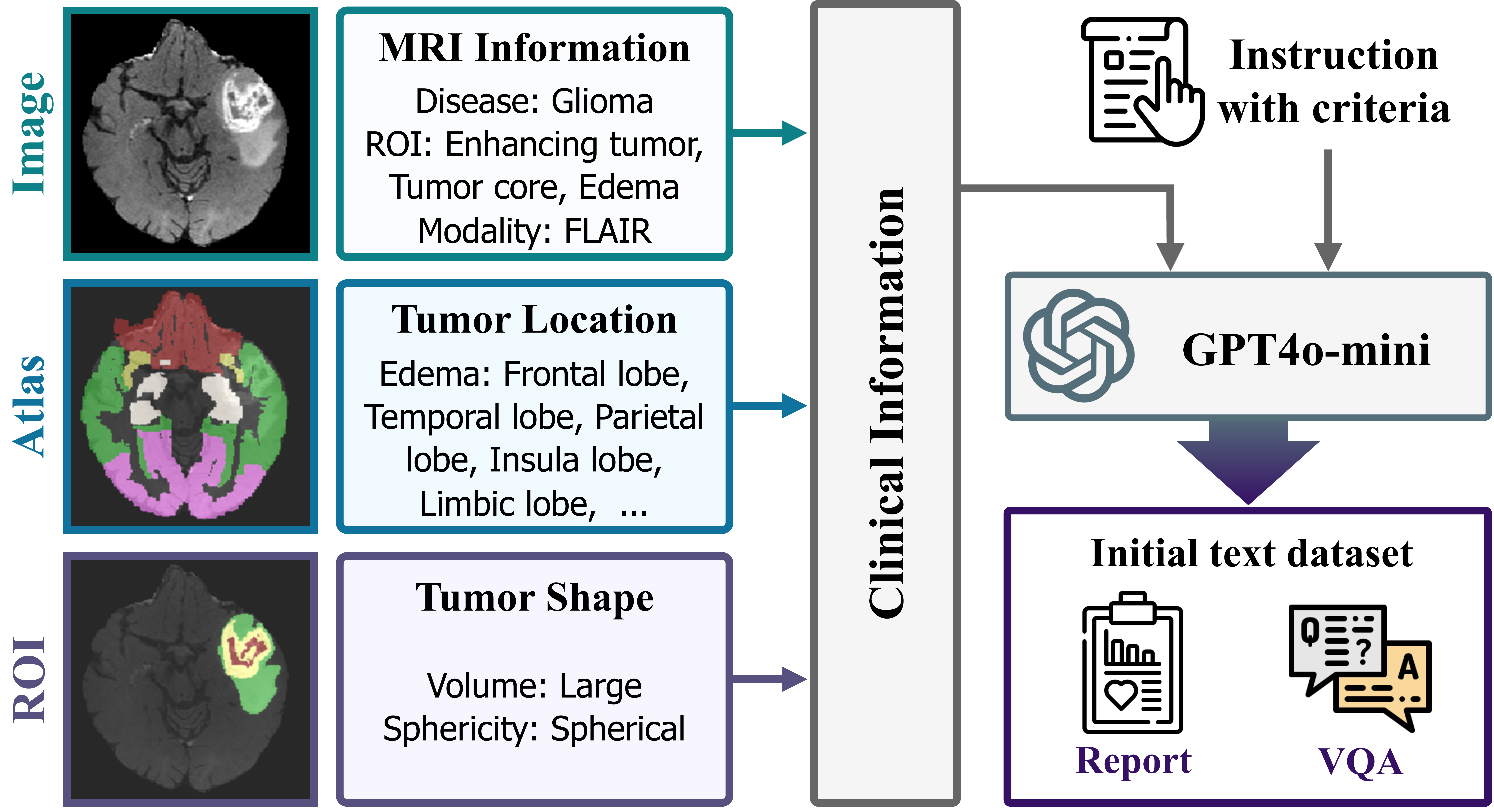}
    \caption{\textbf{Instruction-driven initial report generation}. Reports and VQA data are generated using basic MRI information and tumor location/shape with instructions. The tumor location is provided based on the brain atlas. The combined clinical information with instructions is fed into GPT 4o-mini to generate initial data, such as reports and VQA sets.}
    \label{fig2}
\end{figure}

In summary, our main contributions are as follows:
\begin{itemize}
    \item[1.] We construct a novel 3D image-text paired dataset for brain oncology using open datasets to address the scarcity of publicly available MRI-text datasets.
    \item[2.] We propose a multi-LLM collaborative report generation framework for refinement to improve the quality and reliability of MRI-based medical reports, particularly in scenarios where expert clinicians are unavailable, enabling clinical decision support even under resource-limited conditions.
    \item[3.] We develop a VLM trained on the generated dataset, leveraging a VQ-GAN encoder for efficient multimodal alignment.
\end{itemize}

\section{Method}
In this study, we propose a three-step process to enable a VLM to understand 3D MRI images. Section \ref{sec2.1} presents the process for generating text data (e.g., reports and VQA sets) for datasets that do not contain 3D image-text pairs. Section \ref{sec2.2} describes a multi-LLM collaborative refinement process that uses LLMs to remove inaccurate, unnecessary, and hallucinating information from the generated data. Section \ref{sec2.3} describes the training of our VLM through visual instruction tuning using the generated 3D image-text pairs.

\begin{figure} [t]
    \centering
    \includegraphics[width=0.975\columnwidth]{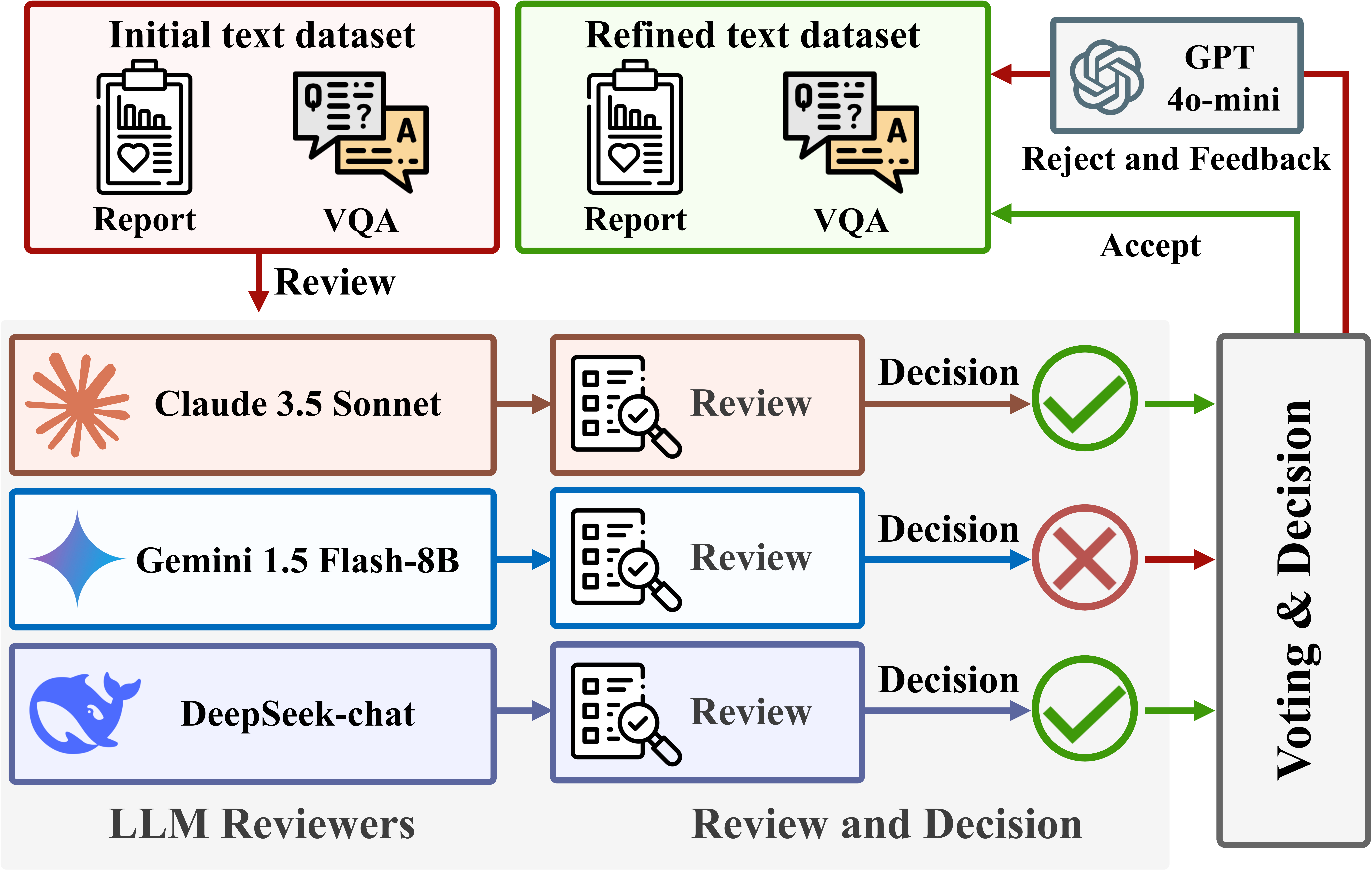}
    \caption{\textbf{Report refinement process with multiple LLM reviewers}. The generated initial text is reviewed by LLM reviewers. The reviewers decide to accept or reject it. If two or more reviewers accept it, the text is added to the refined text dataset (green line flow). If one or fewer accept it, the text is refined using each reviewer’s feedback (red line flow).}
    \label{fig3}
\end{figure}

\begin{figure*} [t]
    \centering
    \includegraphics[width=0.95\textwidth]{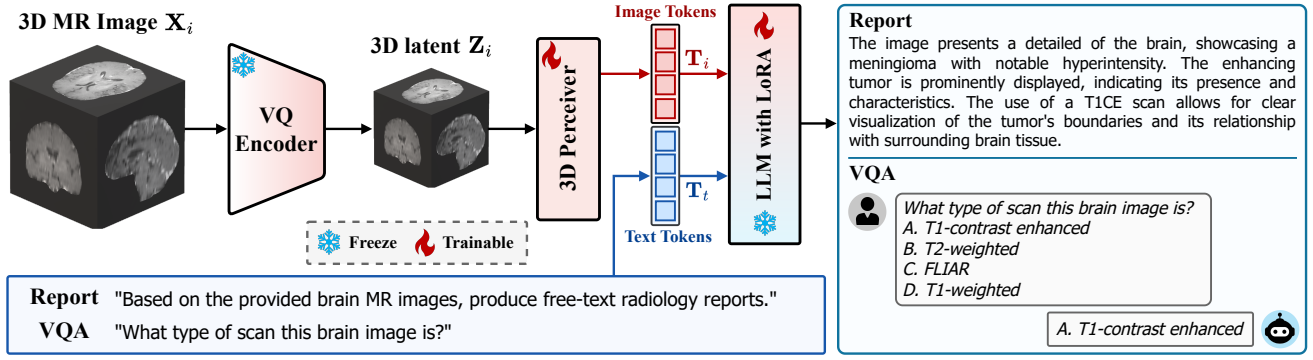}\caption{\textbf{Illustration of training framework with a 3D image token}. A 3D image is compressed into a 3D latent using a pretrained VQ encoder. Then a perceiver converts the latent into an image token, which is used to train an LLM with an instruction. The text token represents the instruction provided during training. The backbone of the LLM is frozen and only LoRA is fine-tuned.}
    \label{fig4}
\end{figure*}

\subsection{Instruction-Driven Data Generation} % with prompt
\label{sec2.1}
3D medical images from open datasets rarely have image-text pairs; therefore, we need to generate data to train the VLM. Recent studies have explored methods of using LLMs to generate text data for training VLMs \cite{liu2023visual,li2023llavamed,cui2024biomedical}; we follow their directions. Grounded in the context of brain oncology, we extract basic image information, such as disease details and modality, along with the location and shape of the tumor. This process is illustrated in Fig. \ref{fig2}. In detail, the tumor location is determined by dividing the brain into seven anatomical regions \cite{hawrylycz2012anatomically,desikan2006automated} and then finding the region that overlaps with the tumor. The tumor shape is measured using predefined radiomics \cite{aerts2014decoding} feature formulas that reflect volume and sphericity, which are then used to classify tumor size. Using this information, we generate a report and a VQA set using an LLM. To avoid errors in the text data, we define an instruction with criteria in advance {shown in Fig \ref{fig2}}. The instructions are used to minimize misinformation and hallucinations in the generated text, ensuring that only accurate information is included. An example of such instructions is as:

\vspace{9pt}
\begin{adjustbox}{scale=0.77}
\begin{minipage}{1.3\linewidth}
\begin{Verbatim}[frame=single, numbersep=5pt,
                 commandchars=\\\{\}]
\textbf{Instruction with criteria}: "You are an radiologist 
specialized in biomedical topics. Please meticulously 
extract all possible visual details, and when gen-
erating questions and answers. Below are requirements 
for generating the questions and answers in the con-
versation: 
- Ensure that questions are diverse and cover a range 
of visual aspects of the image
- Avoid hallucination when generating reports and en-
sure that the generated report is relevant to text 
provided ..."
\end{Verbatim}
\end{minipage}
\end{adjustbox}
\vspace{6pt}

\noindent The LLM may still produce unnecessary, inaccurate, or hallucinatory results; thus, further processing is required to verify and filter the generated text. This issue is addressed in Section \ref{sec2.2}.

\subsection{Report Refinement with LLM Collaboration} % Refinement with Self-Feedback
\label{sec2.2}

Ensuring the accuracy and reliability of the generated medical reports is essential for creating a trustworthy dataset. However, relying on a single LLM for report generation may result in incomplete, inaccurate, or hallucinated content, leading to a compromised dataset. To address this issue, we implement a multi-LLM collaborative report refinement in which multiple independent LLMs review the generated reports and provide feedback for improvement. This validation process enhances the factual correctness and clinical consistency of the generated reports before they are incorporated into the training dataset.

Once an initial report is generated, it is reviewed by three independent LLMs (Claude \cite{TheC3}, Gemini \cite{team2024gemini}, and DeepSeek \cite{bi2024deepseek}), each receiving both the generated report and the original input text used during the initial report generation. The evaluation is based on two primary criteria: accuracy of information integration and consistency with clinical knowledge. The first criterion ensures that all information is included accurately and without distortion, whereas the second verifies alignment with clinical standards to prevent misleading information. Each LLM reviewer independently determines whether the report meets these criteria and classifies it as either acceptable or not.

If at least two of the three reviewing LLMs reject the report, it is returned to the original language model (Chat GPT4o-mini \cite{achiam2023gpt}), along with the given feedback for report generation. The original model then regenerates the report by incorporating this feedback. By integrating this LLM-assisted refinement process, we significantly reduce the incidence of errors related to incorrect medical reasoning and missing attributes. This improvement can be observed in Fig. \ref{fig1}. Consequently, this approach improves the overall quality of the dataset and provides reliable training data for the 3D VLM described in section \ref{sec2.3}.

\begin{figure*} [t]
    \centering
    \includegraphics[width=0.975\textwidth]{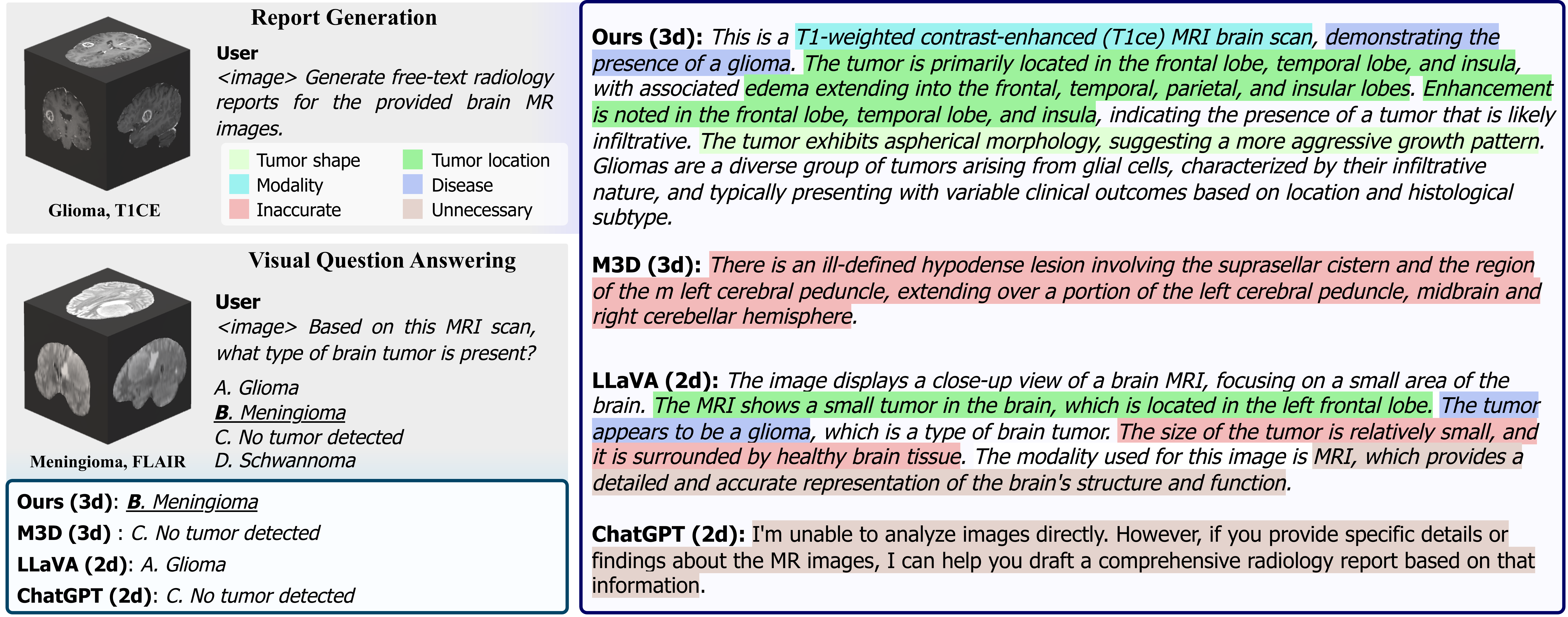}
    \caption{\textbf{Example of text generation of various VLMs}. Left: Results of visual question answering. VQA is closed-ended and each model infers answers based on the input and questions. Right: Results of report generation. Each model generates a report on 3D MRI, highlighting the important and unnecessary elements in the image.}
    \label{fig5}
\end{figure*}

\begin{table*} [t]
    \setlength{\tabcolsep}{7pt} 
    \centering
    \caption{\textbf{Performance of report generation}. Each model is fed an image and instructions, and then generates a report related to the image. We measure the similarities of the generated reports. The 2D VLM is given an axial slice corresponding to the center of the ROI, while the 3D VLM is given a 3D volume. \textbf{Bold} denotes best performance. }
    \label{table1}
    \centering
    \scalebox{1}{
        \begin{tabular}{cl|cccc|cccc}
            \toprule
            \multicolumn{2}{c}{\textbf{Dataset}} &  \multicolumn{4}{|c}{\textbf{BraTS2021-GLI} \cite{baid2021rsna}}  & \multicolumn{4}{|c}{\textbf{BraTS2023-MEN} \cite{labella2023asnrmiccai}} \\
            \cmidrule(lr){1-2} \cmidrule(lr){3-6} \cmidrule(lr){7-10} 
            \textbf{Dim} & \textbf{Model} & \scriptsize \textbf{BLEU} & \scriptsize \textbf{ROUGE} & \scriptsize \textbf{METEOR} & \scriptsize \textbf{BERT-F1} &  \scriptsize \textbf{BLEU} & \scriptsize \textbf{ROUGE} & \scriptsize \textbf{METEOR} & \scriptsize \textbf{BERT-F1}  \\
            \midrule
            \multirow{3}{*}{2D} & LLaVA \cite{liu2023visual} & 0.89 & 18.97 & 12.04 & 69.57 & 0.84 & 18.38 & 11.97 & 69.44  \\
            & LLaVA-Med \cite{li2023llavamed} & 0.46 & 16.3 & 9.99 & 68.47 & 0.55 & 17.19 & 10.79 & 68.9  \\
            & ChatGPT 4o-mini \cite{achiam2023gpt} & 0.81 & 16.90 & 10.22 & 66.31 & 0.74 & 16.75 & 10.31 & 66.22  \\
            \midrule
            \multirow{3}{*}{3D} & M3D (LLaMA2-7B) \cite{bai2024m3d} & 0.16 & 9.65 & 4.94 & 66.17 & 0.21 & 9.58 & 5.05 & 66.30 \\
            & M3D (Phi3-4B) \cite{bai2024m3d} & 0.12 & 8.43 & 3.77 & 65.38 & 0.17 & 9.07 & 4.53 & 65.62 \\
            & \textbf{Ours}     & \textbf{11.72} & \textbf{43.57} & \textbf{33.59} & \textbf{80.61} & \textbf{9.76} & \textbf{42.56} & \textbf{33.18} & \textbf{79.29} \\
            \bottomrule
        \end{tabular}
    }
\end{table*}

\subsection{Visual Instruction Tuning for 3D Medical Imaging} % 3D Visual Instruction Tuning
\label{sec2.3}

To enable the VLM to understand 3D medical images, we employ visual instruction tuning \cite{liu2023visual,bai2024m3d,lai2024e3d} (shown in Fig. \ref{fig4}). This approach uses image-text pairs, including radiology reports and VQA datasets, to fine-tune the model by aligning 3D image and text tokens in a shared multimodal space. First, we encode 3D MRI images into latent representations using a VQ encoder \cite{esser2021taming,kim2024adaptive}. These latent representations are then tokenized into image tokens by the 3D perceiver, which aggregates and projects the high-dimensional image features into a format suitable for input to the LLM. 
\begin{equation}
\textbf{Z}_i = \mathcal{E}_{VQ}(\textbf{X}_i), \ \ \  \textbf{T}_i = \text{Flatten}(\text{Conv3D}(\textbf{Z}_i)),
\end{equation}
where $\textbf{X}_i$ is the 3D image, $\textbf{Z}_i$ is the 3D latent representation, $\textbf{T}_i$ is the image tokens, $\mathcal{E}_{VQ}$ is the pretrained VQ encoder, and Conv3D is the 3D convolution block for the i-th patient. In parallel, corresponding text tokens for the i-th patient from the generated reports and VQA data are fed to the LLM. The image tokens $\textbf{T}_i$ and instruction text tokens $\textbf{T}_t$ are fed into an LLM fine-tuned with Low-Rank Adaptation (LoRA) \cite{hu2021lora}, enabling the model to align multimodal inputs and generate clinically relevant outputs such as detailed reports and VQA responses. This training process allows the VLM to bridge the gap between the visual complexity of 3D medical images and the semantic information of textual descriptions, enabling the generation of high-quality reports and VQA datasets.                                     
\section{Experiments}

\noindent \textbf{Datasets.} We used two datasets, BraTS2021-GLI \cite{baid2021rsna} and BraTS2023-MEN \cite{labella2023asnrmiccai}, to generate the report and VQA datasets, and to train and evaluate our 3D VLM model. BraTS2021-GLI is a dataset with 1251 subjects (glioma, 1125 for training and 126 for testing), and BraTS2023-MEN is a dataset with 1000 subjects (meningioma, 900 for training and 100 for testing). Both datasets include T1, T1ce, T2, and FLAIR modalities, providing a total of 9004 3D volumes. The experimental procedures involving human subjects described in this paper were approved by the Institutional Review Board of Sungkyunkwan University.

\noindent \textbf{Implementation details.}
For report generation, we utilized ChatGPT 4o-mini \cite{achiam2023gpt} to produce the initial reports. These reports were then subjected to a self-refinement process involving three additional LLMs: Deepseek-chat \cite{bi2024deepseek}, Claude 3.5 Sonnet \cite{TheC3}, and Gemini 1.5 Flash-8B \cite{team2024gemini}. Each LLM evaluated the reports based on predefined criteria, providing a pass or fail vote and corresponding feedback. All LLMs were accessed and integrated using their respective APIs.
To train the VLM, we utilized Vicuna-7B-1.5 \cite{chiang2023vicuna} as the baseline LLM and followed a two-stage fine-tuning process. During the first five epochs, only the 3D Perceiver was trained, whereas the LLM parameters were frozen, using the AdamW \cite{loshchilov2018decoupled} optimizer with a learning rate of 1e-6. In the next five epochs, both the 3D Perceiver and the LLM were fine-tuned, with LoRA applied to the LLM. The LoRA configuration was set to \( r = 32 \) and \(\alpha = 32\). All experiments were performed using the PyTorch framework on a single NVIDIA H100 GPU. 
\section{Results}

In this study, all evaluations, including report generation and assessment, were conducted using our previously constructed image-text paired dataset because there are no text-paired 3D MRI datasets. Further, there are currently no established benchmarks or alternative evaluation methods for directly assessing automatically generated reports on these datasets. Therefore, we focus our evaluation on the generated text’s quality and the model’s ability to produce clinically relevant, semantically accurate reports and answers. We assessed whether the model can extract context from 3D MRI, such as modality, disease, and location information, explain relevant information in a text report, and answer questions.

\begin{figure*} [t]
    \centering
    \includegraphics[width=\textwidth]{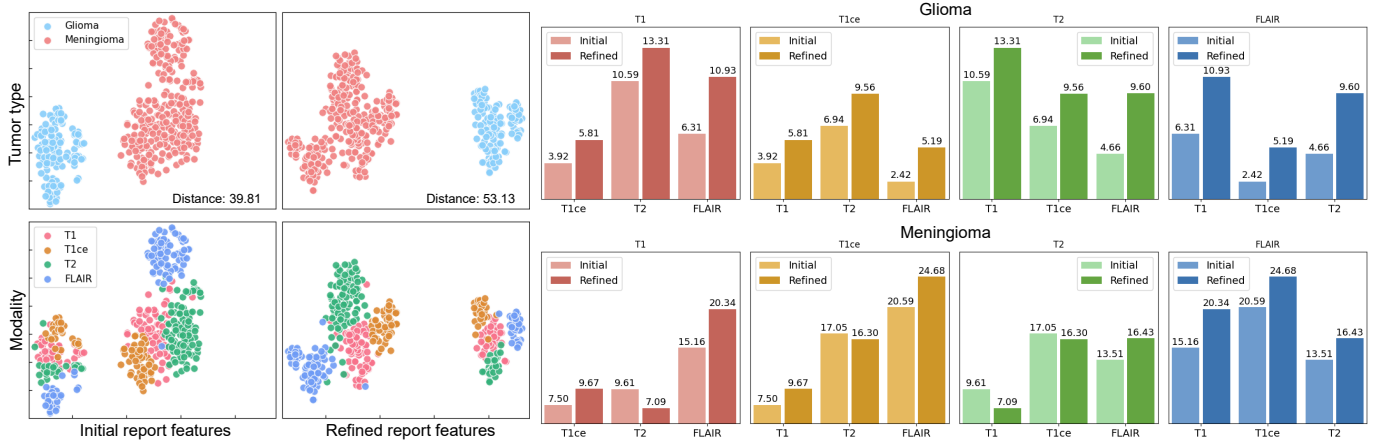}
    \caption{\textbf{Visualization of the text embedding for the initial and refined reports.} Left: Feature space visualization with t-SNE showing clear separation by tumor type and modality. Right: Displays the distances between the centroids of the clusters identified in left.}
    \label{fig6}
\end{figure*}

\begin{table*} [t]
    \setlength{\tabcolsep}{10pt} 
    \centering
    \caption{\textbf{Performance of visual question answering}. Each model receives questions in three categories (tumor type, modality, and tumor location) and is evaluated based on accuracy, F1 score, sensitivity, and precision. \textbf{Bold} denotes best performance.}
    \label{table2}
    \centering
    \scalebox{1}{
        \begin{tabular}{cl|cccc|cccc}
            \toprule
            \multicolumn{2}{c}{\textbf{Dataset}} &  \multicolumn{4}{|c}{\textbf{BraTS2021-GLI} \cite{baid2021rsna}}  & \multicolumn{4}{|c}{\textbf{BraTS2023-MEN} \cite{labella2023asnrmiccai}} \\
            \cmidrule(lr){1-2} \cmidrule(lr){3-6} \cmidrule(lr){7-10} 
            \textbf{Dim} & \textbf{Model} & \textbf{ACC.} & \textbf{F1} & \textbf{Sens.} & \textbf{Prec.} & \textbf{ACC.} & \textbf{F1} & \textbf{Sens.} & \textbf{Prec.}  \\
            \midrule
            \multirow{3}{*}{2D} & LLaVA \cite{liu2023visual} & 41.97 & 27.01 & 26.26 & 31.02 & 23.23 & 26.62 & 26.74 & 29.80   \\
            & LLaVA-Med \cite{li2023llavamed}  & 46.68 & 16.11 & 14.70 & 19.54 & 44.73 & 24.51 & 22.51 & 29.61 \\
            & ChatGPT 4o-mini \cite{achiam2023gpt} & 53.22 & 52.24 & 52.06 & 54.73 & 39.02 & 49.40 & 48.85 & 53.82 \\
            \midrule
            \multirow{3}{*}{3D} & M3D (LLaMA2-7B) \cite{bai2024m3d}  & 50.83 & 35.47 & 34.79 & 37.14 & 29.51 & 23.82 & 23.68 & 24.27 \\
            & M3D (Phi3-4B) \cite{bai2024m3d}& 45.33 & 37.50 & 37.71 & 39.24 & 38.64 & 40.98 & 41.60 & 43.29 \\
            & \textbf{Ours}   & \textbf{91.69} & \textbf{79.39}  & \textbf{79.91} & \textbf{80.03} & \textbf{88.95} & \textbf{84.03}  & \textbf{84.33} & \textbf{84.11} \\
            \bottomrule
        \end{tabular}
    }
\end{table*}

\subsection{Comparing the performance of various VLMs}
\noindent \textbf{Report Generation.}
We compared our report generation method against recent VLM baselines for the text-augmented BraTS2021-GLI and BraTS2023-MEN datasets. We evaluated each method on evaluation metrics for language models (BLEU \cite{papineni2002bleu}, ROUGE \cite{lin2004rouge}, METEOR \cite{banerjee2005meteor}, and BERT-F1 \cite{Zhang2020BERTScore}). For the 2D models (LLaVa and ChatGPT), a single axial slice centered on the tumor was provided as the input. In contrast, M3D used 32 axial slices, also centered on the tumor. Our method used all axial slices. Table \ref{table1} shows that our model outperformed the comparison methods, including LLaVA, LLaVA-Med, ChatGPT 4o-mini, and M3D. Notably, our method achieved the highest BERT-F1 scores on both datasets, indicating that the generated reports were fluent, comprehensive, and semantically well-aligned with the reference texts. These results demonstrate that our method effectively leverages image-text inputs to produce coherent and diagnostically meaningful radiology reports. Furthermore, in Fig. \ref{fig5}, the report generated by our method omits inaccurate and unnecessary information while providing detailed explanations of the brain MRI and tumor, unlike the other comparison models. This is because the 2D models have access to a single axial slice, and thus cannot capture the full 3D context, limiting their image understanding. In contrast, although the M3D (3D model) processes 32 axial slices, it was originally designed for CT and thus is suboptimal for MRI.

\vspace{3pt}
\noindent \textbf{Visual Question Answering.}
We evaluated VQA performance on three categories of questions (tumor type, modality, and tumor location) for various methods. Each question was posed in a multiple-choice format, with four possible answers (Fig. \ref{fig5}). As shown in Table \ref{table2}, we reported the accuracy, F1-score, sensitivity, and precision of each model. Compared to the baseline models, our model achieved better performance. These results indicate that our method effectively captures spatial and contextual information for accurate VQA in brain oncology.

\subsection{Evaluation of the refined report}
% During the multi-LLM review stage, a subset of the initial reports was rejected, prompting further refinement. Out of a total of 9,004 reports, 415 underwent this collaborative refinement process. 
We compared the initial and refined reports to ensure that our multi-LLM collaborative refinement improved the quality. We extracted text embeddings using a sentence transformer \cite{reimers-2019-sentence-bert}, which is widely used to capture semantic similarity. We visualized these embeddings using t-SNE \cite{van2008visualizing} to reduce the features to two dimensions. This is illustrated in Fig. \ref{fig6} (a), in which the ranges of the subplots match. Both the initial and refined reports show that the tumor types are clearly separated in the feature space. In the initial reports, the clusters corresponding to glioma and meningioma showed moderate separation. However, following the refinement process, the clusters were better separated as the centroid distances between the two tumor types increased. 

We compared the distances between the centroids of each cluster in the reduced feature space (Fig. \ref{fig6} (b)). Our results showed that the centroids in the refined report were further from each other than those in the initial report, indicating that the semantic distinctions between reports for different tumor types were better defined. This suggests that the refined reports minimize ambiguity and provide more organized information. We also analyzed the intra-cluster separability of different modalities within each tumor type. Our analysis revealed that the modality clusters were more clearly defined after the initial reports were refined. The increased centroid distances among modalities indicate that the refinement process enhanced the clarity and discrimination of modality-specific details in the reports. These results demonstrate that our multi-LLM collaborative refinement approach can enhance the generated reports. 

% \begin{table} [t]
%     \setlength{\tabcolsep}{7pt} 
%     \centering
%     \caption{\textbf{Quantitative performance of VQA on external dataset}.}
%     \vspace{-3pt}
%     \label{table1}
%     \centering
%     \scalebox{0.8}{
%         \begin{tabular}{cl|cccc|cccc}
%             \toprule
%             \multicolumn{2}{c}{\textbf{Dataset}} &  \multicolumn{4}{|c}{\textbf{UCSF-PDGM} \cite{calabrese2022university}} & \multicolumn{4}{|c}{ \textbf{Meningioma-SEG} \cite{vassantachart2023segmentation}} \\
%             \cmidrule(lr){1-2} \cmidrule(lr){3-6} \cmidrule(lr){7-10} 
%             \textbf{Dim} & \textbf{Model} & \textbf{ACC} & \textbf{F1} & \textbf{Sens} & \textbf{Prec} & \textbf{ACC} & \textbf{F1} & \textbf{Sens} & \textbf{Prec}  \\
%             \midrule
%             \multirow{3}{*}{2D} & LLaVA \cite{liu2023visual} & &  &  & & &  & &   \\
%             & LLaVA-Med \cite{li2023llavamed}  & &  &  & & &  & & \\
%             & ChatGPT 4o-mini \cite{achiam2023gpt} & &  &  & & &  & & \\
%             \midrule
%             \multirow{3}{*}{3D} & M3D (LLaMA2-7B) \cite{bai2024m3d} & &  &  & & &  & & \\
%             & M3D (Phi3-4B) \cite{bai2024m3d}& &  &  & & &  & & \\
%             & \textbf{Ours}   & 98.90 & 82.48 & 82.37 & 82.61 & &  & & \\
%             \bottomrule

%         \end{tabular}
%     }
% \end{table}

\section{Discussion}

% Data Generation in brain oncology

% limitations - lack of comparison models

% In this study, we present a multi-LLM collaborative framework to generate 3D brain oncology MRI image-text datasets, which were  relatively scarce, as well as a 3D VLM that exceeds conventional 2D and 3D methods in brain oncology report generation and VQA tasks. By employing multiple LLM reviewers, our method refines the initial reports, reducing the incidence of inaccuracies and hallucinations. Our VLM exhibited enhanced performance in both free-text radiology report generation and VQA tasks. Although our approach demonstrates promising computational results, a limitation of our study is the absence of clinical validation involving expert radiologists. To address this, future research should incorporate thorough clinical evaluation and validation with domain experts, ensuring the clinical applicability and trustworthiness of the automated reports. Furthermore, while our study integrated multimodal data consisting of MRI images and text, future research should integrate other types of clinically relevant information, such as multi-omics data. Overall, integrating 3D imaging with robust LLM-based methods holds promise for improving clinical support in brain oncology.
In this study, we present a multi-LLM collaborative framework to generate 3D brain oncology MRI image-text datasets, which were  relatively scarce, as well as a 3D VLM that exceeds conventional 2D and 3D methods in brain oncology report generation and VQA tasks. By employing multiple LLM reviewers, our method refines the initial reports, reducing the incidence of inaccuracies and hallucinations. Our VLM exhibited enhanced performance in both free-text radiology report generation and VQA tasks. 

Although our approach demonstrates promising computational results, several limitations should be acknowledged. First, our study lacks clinical validation involving expert radiologists; automated NLP metrics alone cannot fully assess clinical correctness, and expert evaluation is needed to confirm diagnostic relevance. Second, both the training and evaluation data are generated and refined by LLMs, which may introduce a self-consistency bias; performance gains could partially reflect alignment within the synthetic data pipeline rather than independent diagnostic capability. Third, our experiments rely on the BraTS datasets, which are highly curated with consistent labels and well-defined masks; generalization to heterogeneous real-world clinical data with noise, artifacts, and variable acquisition protocols remains to be verified. To address these limitations, future research should incorporate thorough clinical evaluation with domain experts, external validation on real-world hospital data, and integrate other types of clinically relevant information, such as multi-omics data. Overall, integrating 3D imaging with robust LLM-based methods holds promise for improving clinical support in brain oncology.

\section*{Acknowledgment}
% This study was supported by the National Research Foundation of Korea (RS-2024-00408040); the AI Graduate School Support Program (Sungkyunkwan University) (RS-2019-II190421); the ICT Creative Consilience program (IITP-2025-RS-2020-II201821); the Artificial Intelligence Innovation Hub program (RS-2021-II212068); and the SMC-SKKU Future Convergence Research Program Grant (SMO125123).
This study was supported by the SMC-SKKU Future Convergence Research Program Grant (SMO125123). It was also supported by the AI Graduate School Support Program (Sungkyunkwan University) (RS-2019-II190421), the National Research Foundation of Korea (RS-2024-00408040), the ICT Creative Consilience Program Grant (IITP-2026-RS-2020-II201821), and the AI Computing Infrastructure Enhancement (GPU Rental Support) User Support Program (RQT-25-090077), all funded by the Ministry of Science and ICT (MSIT), Republic of Korea.

\bibliographystyle{IEEEtran.bst}
\bibliography{refs}

\end{document}